\documentclass{article} 
\usepackage{iclr2027_conference,times}

\usepackage{amsmath,amsfonts,bm}

\def\eqref#1{equation~\ref{#1}}

\def\1{\bm{1}}

\DeclareMathAlphabet{\mathsfit}{\encodingdefault}{\sfdefault}{m}{sl}
\SetMathAlphabet{\mathsfit}{bold}{\encodingdefault}{\sfdefault}{bx}{n}

\newcommand{\method}[1]{#1}
\newcommand{\ours}{\method{MapMergeLLM}\xspace}

\newcommand{\ie}{{\em i.e.}}

\usepackage{hyperref}
\usepackage{url}

\usepackage{graphicx}
\usepackage{enumitem}
\usepackage{booktabs}
\usepackage{adjustbox}
\usepackage{wrapfig}
\usepackage{makecell}
\usepackage{subcaption}
\usepackage{xspace}
\usepackage{hyperref}
\usepackage{cleveref}
\usepackage{multirow}
\usepackage[table]{xcolor}
\definecolor{groupshade}{RGB}{235,242,250}
\usepackage{etoc}

\title{From Fragments to Global Maps: Learning Vectorized Map Aggregation with Large Language Models}

\author{
Ziwei Li$^{1,2}$\thanks{Equal contribution.} \, Yi-Tang Chen$^{3}$\footnotemark[1] \, Xiaoqi Wang$^{1,2}$\thanks{Project lead.} \, Wenbin He$^{1,2}$ \, Han-Wei Shen$^{3}$ \, Liu Ren$^{1,2}$ \\[0.2em]
$^{1}$Bosch Research North America \\
$^{2}$Bosch Center for Artificial Intelligence (BCAI) \\
$^{3}$The Ohio State University \\[0.2em]
\texttt{\{ziwei.li,xiaoqi.wang,wenbin.he2,liu.ren\}@us.bosch.com} \\
\texttt{\{chen.10522,shen.94\}@osu.edu}
}

\iclrfinalcopy 
\begin{document}

\maketitle

\lhead{Preprint}

\begin{abstract}
Large-scale vectorized HD maps provide structured road information that is essential for perception, localization, and planning in autonomous driving. Constructing such maps requires aggregating noisy, fragmented, and overlapping local predictions collected along a vehicle trajectory into a coherent global map.
Existing aggregation methods typically rely on hand-crafted rules for fragment association and refinement. However, a fixed set of thresholds cannot effectively handle variations in road structures and prediction errors, often requiring detector-specific tuning or manual adjustment. To address this limitation, we propose \ours, a data-driven framework that formulates vectorized map aggregation as conditional sequence generation with a large language model.
Given serialized local vectorized maps, our model directly predicts the aggregated global map polylines. 
To reduce dependence on any particular upstream detector, we train the model on synthetic local maps generated from clean vector maps using corruptions that simulate representative prediction errors.
We further introduce a coordinate tokenizer with geometry-aware pretraining to precisely represent map coordinates. In addition, we propose a line-level association loss that explicitly supervises correspondences between local observations of the same map element. 
Experiments on Argoverse2 and nuScenes using multiple recent upstream detectors demonstrate that \ours substantially outperforms heuristic and optimization-based aggregation baselines without detector-specific retraining.
\end{abstract}

\section{Introduction}

\begin{figure}[tb]
  \centering
  \includegraphics[width=1\textwidth]{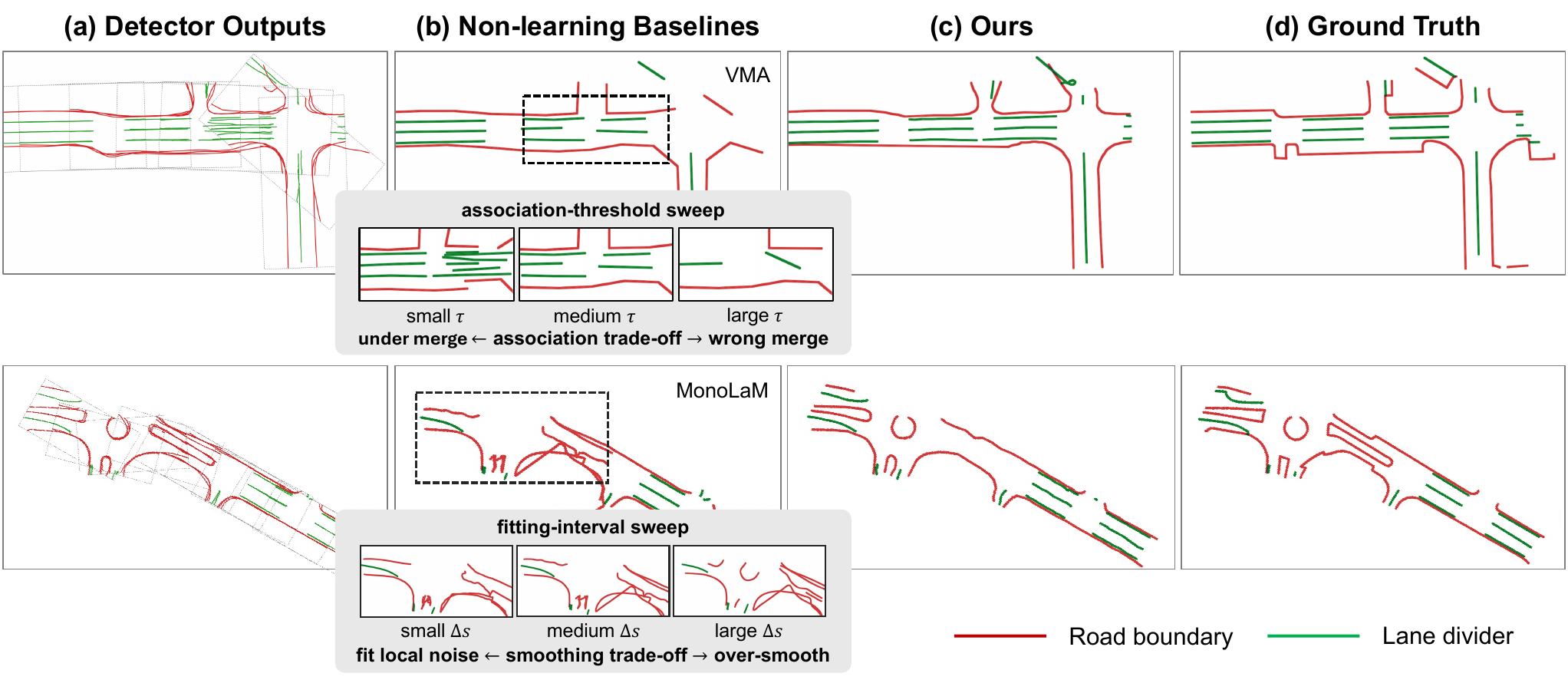}
  \caption{\small \textbf{Non-learning-based aggregation presents association and smoothing trade-offs.} (a) StreamMapNet~\citep{StreamMapNet} predictions accumulated across frames. (b) For the heuristic-based method, a small association threshold ($\tau$) leaves related fragments unmerged, whereas a large threshold incorrectly merges neighboring lanes. For the geometry-based method, a small fitting interval ($\Delta s$) fits local prediction noise, whereas a large interval over-smooths line structures. (c) \ours learns aggregation from data and produces maps close to the (d) ground truth without hand-crafted parameter tuning.}
  \label{fig:teaser}
\end{figure}

High-definition (HD) maps provide detailed information about road structures to support autonomous driving systems~\citep{HDMapNet, MapTRv2}. In vectorized HD maps, line-shaped elements such as lane dividers and road boundaries are represented as polylines, with each polyline represented by an ordered sequence of points.
Recent advances in vectorized map construction enable models to predict these map elements directly from sensor observations~\citep{VectorMapNet, BeMapNet, StreamMapNet, PivotNet}.
As onboard sensors only capture a limited region around the ego vehicle, a large-scale HD map cannot be constructed from a single observation, and instead, it must be built from local vectorized predictions collected along a driving sequence. 
\par
However, simply accumulating these predictions results in redundant, fragmented, and geometrically inconsistent map elements, as the same road structures are repeatedly observed across overlapping local regions. We refer to the process of associating, merging, and refining these local predictions into a coherent global map as \textit{vectorized map aggregation}.
This task is challenging because errors in local map predictions, together with overlapping observations, make it difficult to correctly associate fragments of the same map structure and accurately recover its underlying geometry.
\par
Despite its importance, only a few studies have addressed this problem. Existing methods typically rely on hand-crafted geometric rules with fixed parameters, such as distance thresholds, overlap criteria, or sampling intervals~\citep{VMA, MonoLaM}, which may not generalize well across diverse road structures or prediction errors from different upstream detectors.
For example, in scenes with dense parallel lanes, noisy observations of one lane may be difficult to distinguish from observations of neighboring lanes. Consider a distance-based association threshold $\tau$, where two fragments are considered for merging if their geometric discrepancy falls below $\tau$. As shown in the top zoomed-in views of Fig.~\ref{fig:teaser}(b), a larger $\tau$ may incorrectly merge neighboring lanes, whereas a smaller threshold better preserves their distinction but leaves related fragments of the same lane unmerged.
Geometric refinement presents a similar trade-off. As shown in the bottom zoomed-in views of Fig.~\ref{fig:teaser}(b), geometry-based methods~\citep{cheng2021road, MonoLaM} often rely on a fixed fitting interval $\Delta s$ that controls how closely the refined line follows local observations. A smaller $\Delta s$ allows the line to capture finer geometric variations but also makes it more sensitive to prediction noise, whereas a larger $\Delta s$ provides stronger smoothing but may drop important geometric details. In the illustrated example, a larger $\Delta s$ better recovers the circular curb but over-smooths other curb structures.
These examples show that no single parameter setting is suitable for all road geometry and detection error patterns. The appropriate association threshold or fitting interval depends on the local context, such as lane spacing, curvature, and detector noise patterns, and may vary even within the same scene. This motivates us to learn the aggregation process from data, enabling fragment association and geometric refinement to adapt to the observations rather than being controlled by globally fixed hand-crafted rules.
\par
Therefore, we propose \textbf{\ours}, which formulates vectorized map aggregation as a conditional sequence generation problem and uses a large language model (LLM)~\citep{minaee2024large, qwen3.5} as the sequence-modeling backbone. 
Given local vectorized maps from multiple frames, the model directly generates the aggregated global polylines from serialized local fragments. However, training such a model presents two key challenges. 
First, the model should generalize across different upstream detectors rather than overfit to the error distribution of a particular one. To achieve this, we construct synthetic training data by applying structured corruptions to clean vector maps, simulating common map detection errors such as duplication, fragmentation, and geometric perturbations. 
Second, as LLMs were originally designed for text generation, adapting them to map generation requires a precise representation of map coordinates. Unlike vanilla tokenization, which may split a numerical value into multiple subword units, we encode each coordinate value as a single token, allowing the model to predict a complete value in a single step.
To improve map generation, we propose geometry-aware pretraining that teaches the model numerical and spatial relationships among the new coordinate tokens. We further introduce a line-level association loss that explicitly supervises correspondences between local fragments, helping the model associate observations of the same map element and reduce duplication. Together, these components enable accurate and reliable global map aggregation across different upstream detectors.
\par
To demonstrate the effectiveness of \ours, we evaluate it on local maps produced by recent vectorized map detectors on the Argoverse2~\citep{Argoverse2} and nuScenes~\citep{caesar2020nuscenes} datasets. 
The experimental results show that \ours consistently outperforms heuristic and optimization-based aggregation baselines, particularly on road boundaries, where it achieves mean relative F1 improvements of 23.1\% on Argoverse2 and 21.9\% on nuScenes over the strongest baseline.
In summary, our main contributions include:
\begin{itemize}
    \item We identify key limitations of existing heuristic and geometry-based map aggregation methods, showing that fixed hand-crafted merging rules struggle to generalize across diverse road structures and detector noise patterns.
    
    
    \item We propose \ours, an LLM-based, data-driven framework for vectorized map aggregation, and introduce coordinate-token pretraining and line-level association loss to improve map generation, supported by detector-agnostic synthetic training data.
    
    \item We conduct comprehensive experiments on Argoverse2 and nuScenes, demonstrating that \ours substantially improves over existing aggregation baselines across multiple recent upstream map detectors.
\end{itemize}

\section{Related Work}

\paragraph{Vectorized HD Map Construction and Aggregation.}
Online vectorized HD map construction aims to recover structured map elements, such as lane dividers and road boundaries, within a local region surrounding the ego vehicle from onboard sensor observations~\citep{HDMapNet, VectorMapNet}. Early approaches, such as HDMapNet~\citep{HDMapNet}, first predict map elements in the BEV space and then vectorize them, while more recent methods directly predict vectorized maps~\citep{VectorMapNet, MapTR, MapTRv2}. Building on this direction, temporal methods further leverage historical observations~\citep{StreamMapNet, chen2024maptracker} or a continuously updated global map prior~\citep{GlobalMapNet} to improve prediction consistency. Beyond local map prediction, several works focus on aggregating local vectorized predictions into global maps. VMA~\citep{VMA} associates and merges map elements using predefined geometric rules, while MonoLaM~\citep{MonoLaM} combines geometric association with graph optimization.
GNMap~\citep{fan2024neural} introduces a learning-based fusion framework, but it rasterizes vectorized map tiles before fusion and outputs a semantic map rather than the global polylines required in our task. These aggregation approaches either rely on hand-crafted geometric rules or involve rasterization. In contrast, we propose a data-driven framework that learns to aggregate fragmented observations directly in the vector domain through conditional sequence generation.

\paragraph{Language Models for Map Generation.}
The sequence modeling capabilities of LLMs have recently been explored for structured geometric prediction, including lane and map generation. \citet{zhang2026ladel} employs an MLLM with lane-specific tokens to predict lane geometry from visual observations. \citet{UniMapGen} formulates vectorized map construction as autoregressive generation, progressively producing map elements from multi-modal inputs and previously constructed map states. \citet{PAMR} further extends this paradigm to persistent mapping by jointly modeling lane geometry, traffic rules, and temporal observations. 
Another line of work explores the use of vision-language reasoning for map refinement. \citet{MapAgent} uses such reasoning to identify and correct errors in predicted lane-level maps. While these studies demonstrate the potential of language models for structured map generation and refinement, their application to aggregating multiple vectorized map observations remains largely unexplored.

\section{Method}
\label{sec:method}
\begin{figure}[thb]
  \centering
  \includegraphics[width=1\textwidth]{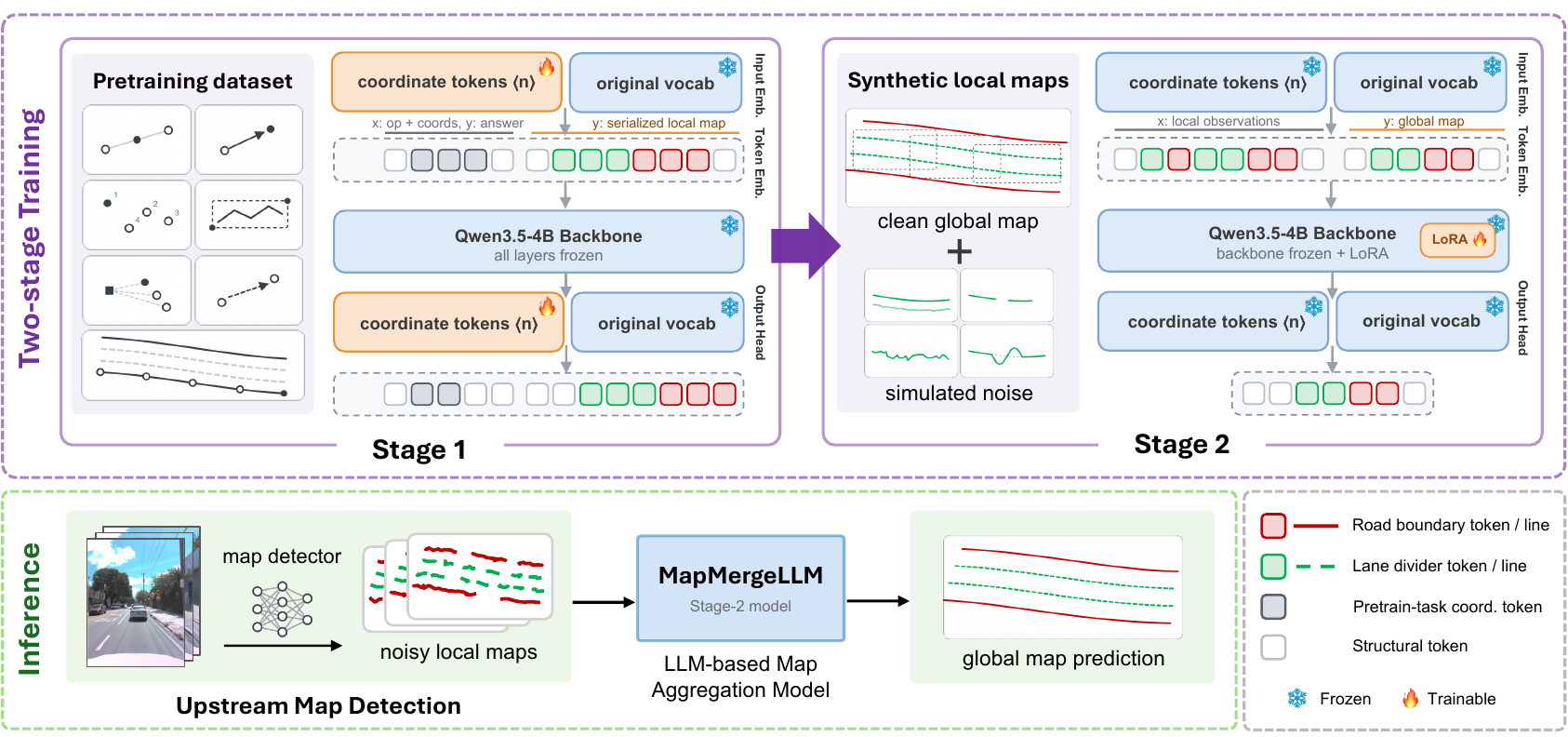}
  \caption{\small \textbf{Overview of \ours}. Training consists of two stages: Stage 1 performs coordinate-token pretraining to learn representations for the newly introduced coordinate tokens, and Stage 2 performs LoRA-based fine-tuning for map aggregation using synthetic noisy local maps. At inference time, the trained model aggregates local predictions from an upstream detector into a global vectorized map.}
  \label{fig:method_overview}
\end{figure}

We propose \ours, a data-driven framework that formulates vectorized map aggregation as a sequence-to-sequence generation task (Section~\ref{method:problem_formulation}). To avoid depending on detector-specific error patterns, we construct synthetic training pairs by applying structured corruptions to clean vector maps (Section~\ref{sec:synthetic_data}). We further introduce coordinate tokenization and a two-stage training scheme (Section~\ref{method:llm_map_aggregation}), including geometry-aware pretraining followed by map aggregation fine-tuning. A line-level association loss complements token-level supervision by encouraging association of related line fragments and reducing duplicate map predictions. 
At inference time, the model takes serialized local predictions in a common coordinate frame as input and generates the global map polylines. An overview of our LLM-based map aggregation framework is shown in Fig.~\ref{fig:method_overview}.

\subsection{Problem Formulation}
\label{method:problem_formulation}

\paragraph{Vectorized Map Definition.}
We represent each map element by a category $c_k$ and an ordered polyline
$\mathbf{P}_k=(\mathbf{p}_{k,1},\ldots,\mathbf{p}_{k,N_k})$,
where $\mathbf{p}_{k,i}=(x_{k,i},y_{k,i})\in\mathbb{R}^2$.
A vectorized map with $K$ elements is represented as
$\mathcal{M}=[(\mathbf{P}_1,c_1),\ldots,(\mathbf{P}_K,c_K)]$,
where both $K$ and the number of sampled points $N_k$ may vary across maps.

\paragraph{Global Map Aggregation.}
Let $\mathcal{M}^{*}$ denote the clean global map over the region covered by a sequence of local observations, and let $\mathcal{C}_t(\mathcal{M}^{*})$ denote the local submap captured at time $t$. We assume that all local predictions can be accurately aligned using ego poses and focus on element-level prediction errors rather than pose-induced misalignment. Given $L$ local vectorized observations
$\widetilde{\mathcal{M}}^{(1)},\ldots,\widetilde{\mathcal{M}}^{(L)}$,
we formulate map aggregation as
\begin{equation}
\widehat{\mathcal{M}}
=
F_{\theta}
\left(
\widetilde{\mathcal{M}}^{(1)},
\ldots,
\widetilde{\mathcal{M}}^{(L)}
\right),
\end{equation}
where $F_{\theta}$ aggregates redundant and complementary geometric evidence across overlapping observations to recover the global clean map $\mathcal{M}^{*}$. 
\par
Different upstream detectors may produce different error patterns due to differences in model architecture and sensing modality.
Rather than training $F_{\theta}$ on predictions from a particular detector, we synthesize local observations from a detector-agnostic corruption distribution,
\begin{equation}
\widetilde{\mathcal{M}}^{(t)}
\sim
q_{\phi}
\left(
\cdot\mid\mathcal{C}_t(\mathcal{M}^{*})
\right),
\end{equation}
where $q_{\phi}$ captures representative prediction errors shared across detectors.
At inference time, $F_{\theta}$ is applied directly to local predictions from different upstream detectors.
The construction of $q_{\phi}$ is described in the following section.

\subsection{Detector-Agnostic Corruption Process}
\label{sec:synthetic_data}
We design the detector-agnostic corruption distribution $q_{\phi}$ to model \textit{recoverable} prediction errors, \ie, errors that can be corrected by aggregating vectorized local observations without access to the original sensor data.
Although individual local predictions may be fragmented, duplicated, or geometrically distorted, the overlapping observations should collectively retain sufficient evidence to recover the underlying map element. Therefore, we focus on four representative failure modes: instance duplication, endpoint corruption and fragmentation, unstructured point-wise perturbation, and structured local geometric perturbation.
\par
Let $T_{\mathrm{dup}}$, $T_{\mathrm{frag}}$, $T_{\mathrm{jit}}$, and $T_{\mathrm{str}}$ denote the four polyline-level corruption operators. Each operator acts on a single polyline and may retain, replace, split, or duplicate the corresponding instance. We lift each operator $T$ to a map-level operator $\mathcal{T}$ by independently applying it to each polyline with probability $\pi_T$, while leaving unselected polylines unchanged. If an operator produces multiple instances, all resulting polylines are retained in the map collection.
We then apply the four map-level operators sequentially to generate a synthetic noisy local map as:
\begin{equation}
    \widetilde{\mathcal{M}}^{(t)}
    =
    \mathcal{T}_{\mathrm{str}}
    \circ
    \mathcal{T}_{\mathrm{jit}}
    \circ
    \mathcal{T}_{\mathrm{frag}}
    \circ
    \mathcal{T}_{\mathrm{dup}}
    \left(
        \mathcal{C}_t(\mathcal{M}^{*})
    \right).
\end{equation}
The operators are applied sequentially, allowing multiple failure modes to coexist within a single local map. The parameters of each operator are sampled from predefined ranges, and the resulting composition defines the corruption distribution $q_{\phi}$.

\paragraph{Instance Duplication.}
Duplicate detections occur when multiple predicted instances correspond to the same underlying map element. For a selected polyline $\mathbf{P}$, we retain the original instance and generate an additional copy $\mathbf{P}^{\mathrm{dup}}$. The duplicate is resampled along arc length with a randomly varied sampling density, and a subset of its points is perturbed with small zero-mean Gaussian noise to simulate slight geometric differences between duplicate predictions. The duplication operator is defined as
\begin{equation}
    T_{\mathrm{dup}}(\mathbf{P})
    =
    [\mathbf{P},\mathbf{P}^{\mathrm{dup}}].
\end{equation}

\paragraph{Endpoint Corruption and Fragmentation.}
Local map predictions may contain truncated endpoints or fragmented map elements.
For a selected polyline $\mathbf{P}$, we first perform arc-length resampling and randomly choose an endpoint $e\in \{\mathrm{head},\mathrm{tail}\}$ and a corruption mode $m\in \{\mathrm{trunc},\mathrm{split}\}$. The corruption operator is defined as
\begin{equation}
    T_{\mathrm{frag}}(\mathbf{P}; e,m)
    =
    \mathcal{O}_{m}(\mathbf{P},e),
\end{equation}
where $\mathcal{O}_{\mathrm{trunc}}$ removes a short portion near the selected endpoint, while $\mathcal{O}_{\mathrm{split}}$ partitions the polyline near that endpoint into two fragments. Both operations preserve most of the underlying geometry while simulating incomplete and fragmented predictions.

\paragraph{Unstructured Point-wise Perturbation.}
Even when a map element is correctly detected, its predicted geometry may contain sparse local deviations due to sensor noise or prediction uncertainty. For a selected polyline $\mathbf{P}$, we randomly select a subset of sampled points and independently perturb their coordinates with zero-mean Gaussian noise:
\begin{equation}
    \mathbf{p}^{\mathrm{jit}}_i
    =
    \mathbf{p}_i
    +
    \boldsymbol{\epsilon}_i,
    \qquad
    \boldsymbol{\epsilon}_i
    \sim
    \mathcal{N}
    \left(
    \mathbf{0},
    \sigma_{\mathrm{jit}}^2\mathbf{I}
    \right),
\end{equation}
while leaving all other points unchanged. We denote the resulting perturbed polyline as
$\mathbf{P}^{\mathrm{jit}} = T_{\mathrm{jit}}(\mathbf{P})$.
This operator simulates spatially uncorrelated point-wise errors without introducing coherent deformation of the polyline.

\paragraph{Structured Local Geometric Perturbation.}
In addition to independent point-wise errors, local map predictions may present smooth, spatially correlated geometric deviations. For example, in regions with closely spaced or complex road markings, a detector may be influenced by nearby lane markings or road boundaries, producing smooth local deviations in the predicted geometry. To simulate such errors, we select a local interval along a polyline and apply a sinusoidal displacement along its local normal direction:
\begin{equation}
    \mathbf{p}^{\mathrm{str}}_i
    =
    \mathbf{p}_i
    +
    \mathbb{I}[s_i \in [s_a,s_b]]
    A
    \sin
    \left(
    \frac{2\pi(s_i-s_a)}{\lambda}
    \right)
    \mathbf{n}_i,
\end{equation}
where $s_i$ denotes the cumulative arc-length position, $\mathbf{n}_i$ is the local unit normal, $A$ controls the displacement amplitude, and $\lambda$ denotes the wavelength. The corrupted interval $[s_a, s_b]$ is centered along the polyline, with length $s_b - s_a = n_h \lambda / 2$, where $n_h$ is an integer specifying the number of sinusoidal half-periods. This construction ensures zero displacement at both interval boundaries. We denote the resulting polyline as $\mathbf{P}^{\mathrm{str}} = T_{\mathrm{str}}(\mathbf{P})$.



\subsection{LLM-Based Map Aggregation}
\label{method:llm_map_aggregation}

\paragraph{Model Architecture.}
Given the local vectorized observations defined above, we serialize each map as a sequence of map elements, where each element is represented by its category and the ordered coordinates of its polyline. 
Standard subword tokenizers, such as byte-pair encoding (BPE)~\citep{sennrich2016neural}, typically construct tokens based on how frequently character sequences appear in text, without explicitly considering their numerical meaning. As a result, a single coordinate value may be split into multiple subword units~\citep{kreitner2026bittokens, singh2024tokenization}. 
For map aggregation, the model must compare coordinates across observations and generate corrected coordinates for polyline merging and refinement. Such fragmentation makes precise coordinate generation more difficult, as every token of a coordinate value must be predicted correctly. An error in any token can substantially change the resulting map position.
\par
Therefore, we use \textit{coordinate tokenization}, which encodes each discretized coordinate value as a single new token. Given a point $p_i=(x_i,y_i)$ and a spatial resolution $\delta$ in meters, we quantize each coordinate to an integer:
\begin{equation}
    \tilde{z}_i
    =
    \operatorname{round}\left(\frac{z_i}{\delta}\right),
    \qquad
    z_i \in \{x_i,y_i\}.
\end{equation}
For example, a coordinate value of $12.34~\mathrm{m}$ is mapped to $123$ at $\delta=0.1~\mathrm{m}$. We then extend the vocabulary with a new token $\langle n\rangle$ for each integer value within the supported coordinate range:
\begin{equation}
    \mathcal{V}_{\mathrm{coord}}
    =
    \left\{
    \langle n\rangle
    \;\middle|\;
    n\in\mathbb{Z},
    \;
    \tilde{z}_{\min}
    \le n \le
    \tilde{z}_{\max}
    \right\}.
\end{equation}
During decoding, each predicted coordinate token $\langle n\rangle$ is converted back to meters as $z=n\delta$.
\par
\ours employs Qwen3.5-4B~\citep{qwen3.5} as its backbone. We add the coordinate tokens to its vocabulary and expand the input embedding matrix and language modeling (LM) head with corresponding token-specific rows. For parameter-efficient adaptation, we apply LoRA~\citep{hu2022lora} to the selected linear layers of the backbone. The training procedure consists of two stages, as described below.

\paragraph{Stage 1: Coordinate-Token Pretraining.}
Although each new coordinate token represents a complete coordinate value, the model still needs to learn its numerical meaning and how to use it in map sequences. Map aggregation requires comparing coordinates across observations and adjusting them to recover the underlying geometry. Therefore, we propose a \textit{coordinate-token pretraining} scheme that uses geometric and arithmetic tasks, such as 2D coordinate translation, midpoint prediction, and point sorting, to help the model learn how to compute and compare coordinates before the model is trained for map aggregation. We also include next-token prediction on serialized local maps to help the model learn how to use the new tokens within polyline sequences. Details of the pretraining tasks are provided in Appendix~\ref{appendix:pretraining_tasks}.
\par
During this stage, we optimize only the newly added coordinate-token rows in the input embedding matrix and LM head, while keeping all original pretrained parameters frozen. 
Let $\theta_0$ denote all frozen pretrained model parameters and $\mathbf{E}_c$ denote the newly added coordinate-token rows in the input embedding matrix and LM head. We optimize $\mathbf{E}_c$ using the following autoregressive cross-entropy loss:
\begin{equation}
    \mathcal{L}_{\mathrm{pre}}(\mathbf{E}_c)
    =
    -\,\mathbb{E}_{(\mathbf{x},\mathbf{y})\sim\mathcal{D}_{\mathrm{pre}}}
    \left[
        \frac{1}{|\mathbf{y}|}
        \sum_{j=1}^{|\mathbf{y}|}
        \log
        p_{\theta_0,\mathbf{E}_c}
        \left(
            y_j
            \mid
            \mathbf{y}_{<j},
            \mathbf{x}
        \right)
    \right].
\end{equation}
Here, $\mathcal{D}_{\mathrm{pre}}$ represents the pretraining dataset. For geometric and arithmetic tasks, $\mathbf{x}$ specifies the operation and input coordinates, and $\mathbf{y}$ is the corresponding answer, with the loss applied only to the answer tokens. For map sequence prediction, we train the model to predict each token in a serialized local map given the previous tokens, and the loss is computed over all tokens.

\paragraph{Stage 2: Map Aggregation Training.}
In Stage 2, we adapt the model for map aggregation by only optimizing the LoRA parameters, while keeping the pretrained model parameters $\theta_0$ and the coordinate-token parameters $\mathbf{E}_c$ learned in Stage 1 frozen, as shown in Fig.~\ref{fig:method_overview}. Each training pair consists of the synthetic noisy local observations $\mathbf{x}$, generated by the corruption process introduced in Section~\ref{sec:synthetic_data}, and the clean global map $\mathbf{y}$ as the target.
We supervise autoregressive map generation with a token-level cross-entropy loss over the target sequence:
\begin{equation}
    \mathcal{L}_{\mathrm{agg}}(\Delta)
    =
    -\,\mathbb{E}_{(\mathbf{x},\mathbf{y})\sim\mathcal{D}_{\mathrm{agg}}}
    \left[
        \frac{1}{|\mathbf{y}|}
        \sum_{j=1}^{|\mathbf{y}|}
        \log
        p_{\theta_0,\Delta,\mathbf{E}_c}
        \left(
            y_j
            \mid
            \mathbf{y}_{<j},
            \mathbf{x}
        \right)
    \right],
\end{equation}
where $\mathcal{D}_{\mathrm{agg}}$ denotes the map aggregation training dataset, and $\Delta$ represents the trainable LoRA parameters. Under teacher forcing, each target token is predicted conditioned on the local observations and the ground-truth prefix $\mathbf{y}_{<j}$.

\paragraph{Line-Level Association Loss.}
When combining two local maps, a natural first step is to determine which line fragments belong to the same underlying map element and should be merged together. Token-level cross-entropy supervises the generated map sequence but does not explicitly supervise these fragment-level correspondences. Therefore, we introduce a \textit{line-level association loss} to encourage fragments of the same map element to share similar representations while separating the unrelated ones, facilitating both fragment association and deduplication. We formulate this objective as a contrastive loss. For each input polyline, we construct a line embedding by mean-pooling the final-layer hidden states of its coordinate tokens:
\begin{equation}
    \mathbf{h}_i
    =
    \frac{1}{|\mathcal{J}_i|}
    \sum_{j\in\mathcal{J}_i}
    \mathbf{h}_j^{\mathrm{token}},
\end{equation}
where $\mathcal{J}_i$ denotes the set of coordinate-token positions belonging to the \(i\)-th local polyline, excluding structural tokens such as commas and parentheses.
\par
Each polyline $i$ is associated with a category $c_i$, a ground-truth global element ID $g_i$, and a frame index $t_i$. We define positive samples as observations of the same global element from adjacent local frames: 
$
    \mathcal{P}(i)
    =
    \left\{
    k\neq i
    \mid
    (c_k,g_k)=(c_i,g_i),
    \;
    |t_k-t_i|=1
    \right\},
$
and negative samples as observations of different global elements: 
$
    \mathcal{N}(i)
    =
    \left\{
    k
    \mid
    (c_k,g_k)\neq(c_i,g_i)
    \right\}.
$
The line-level contrastive objective is defined as: 
\begin{equation}
    \mathcal{L}_{\mathrm{line}}
    =
    -\frac{1}{|\mathcal{A}|}
    \sum_{i\in\mathcal{A}}
    \frac{1}{|\mathcal{P}(i)|}
    \sum_{p\in\mathcal{P}(i)}
    \log
    \frac{
        \exp(\mathbf{z}_i^{\top}\mathbf{z}_p/\tau)
    }{
        \sum_{k\in\mathcal{P}(i)\cup\mathcal{N}(i)}
        \exp(\mathbf{z}_i^{\top}\mathbf{z}_k/\tau)
    },
\end{equation}
where $\mathcal{A}=\{i\mid\mathcal{P}(i)\neq\emptyset\}$ is the index set of input polylines with at least one positive sample, $\mathbf{z}_i=\mathbf{h}_i/\lVert\mathbf{h}_i\rVert_2$, and $\tau$ is the temperature parameter.
The training objective of Stage 2 combines both the token-level generation loss and the line-level association loss:
\begin{equation}
    \mathcal{L}
    =
    \mathcal{L}_{\mathrm{agg}}
    +
    \lambda_{\mathrm{line}} \mathcal{L}_{\mathrm{line}},
\end{equation}
where $\lambda_{\mathrm{line}}$ is a weighting coefficient to balance the two objectives.

\paragraph{Inference.}
At inference time, local map predictions from an upstream detector are transformed into a common coordinate frame and serialized using the proposed coordinate tokenization. Conditioned on the serialized observations, the model autoregressively generates the global map sequence. The generated sequence is then parsed into polylines, with coordinate tokens converted back to coordinate values at resolution \(\delta\).

\section{Experiments}
\label{sec:experiments}

\subsection{Experimental Setup}
\label{sec:exp_setup}

\paragraph{Datasets.}
We evaluate \ours on the Argoverse2~\citep{Argoverse2} and nuScenes~\citep{caesar2020nuscenes} datasets, focusing on lane dividers and road boundaries.
To construct training pairs, we first crop local maps from the ground-truth vectorized maps at intervals of $15\mathrm{m}$ along the ego trajectory, with each crop covering a $60\mathrm{m}\times30\mathrm{m}$ region centered at the corresponding ego position.
We then apply the synthetic corruption process described in Section~\ref{sec:synthetic_data} to these local maps to generate the noisy input observations.
The corresponding target global map is defined over the union of these local regions.
At inference time, we obtain local map predictions using the official checkpoints of MapTracker~\citep{chen2024maptracker} and StreamMapNet~\citep{StreamMapNet}. We first transform these predictions from their respective ego coordinate frames into a common global coordinate frame and then use the aligned predictions as input to our model for aggregation.

\paragraph{Baselines.}
We compare \ours with two representative map aggregation baselines: (1) VMA~\citep{VMA} incrementally merges local vectorized maps using predefined association and merging rules, such as checking whether two polylines sufficiently overlap or whether the endpoints of each polyline are close to the other. (2) MonoLaM~\citep{MonoLaM} represents line elements using Catmull-Rom splines. It associates local observations with existing map elements based on geometric criteria and incrementally updates the map through graph optimization.
We use the default settings of MonoLaM and adjust the parameters of VMA to better preserve the geometry of the predicted lines.


\paragraph{Evaluation Metrics.}
We adopt the line-level matching criterion from MonoLaM to evaluate aggregated global maps over complete driving scenes. Following MonoLaM, a predicted polyline is considered a match to a ground-truth polyline if more than 75\% of the points from the ground-truth polyline are within a distance threshold ($d$) of the prediction. We use greedy matching, where each predicted polyline can be counted as a true positive at most once. We report precision, recall, and F1 scores separately for lane dividers and road boundaries. For each distance threshold, these metrics are computed from counts accumulated over all test scenes. The final results are averaged across $d$ $\in$ $\{0.5\mathrm{m},1.0\mathrm{m},1.5\mathrm{m}\}$.
\par
Implementation and training details are provided in Appendix \ref{appendix:implementation_details}.

\subsection{Results}
\label{sec:results}

\begin{table}[b]
\centering
\caption{\small \textbf{Comparison with map aggregation baselines} using per-frame predictions from MapTracker and StreamMapNet on AV2 and nuScenes datasets. Precision, recall, and F1 (\%) are reported for lane dividers and road boundaries and averaged over matching thresholds $d\in \{0.5,1.0,1.5\}$ meters.}
\label{tab:main_results}
\footnotesize
\setlength{\tabcolsep}{3pt}
\begin{tabular}{l cccccc cccccc}
\toprule
& \multicolumn{6}{c}{Argoverse2} & \multicolumn{6}{c}{nuScenes} \\
\cmidrule(lr){2-7}\cmidrule(lr){8-13}
& \multicolumn{3}{c}{Lane Divider} & \multicolumn{3}{c}{Road Boundary}
& \multicolumn{3}{c}{Lane Divider} & \multicolumn{3}{c}{Road Boundary} \\
\cmidrule(lr){2-4}\cmidrule(lr){5-7}\cmidrule(lr){8-10}\cmidrule(lr){11-13}
Method & Prec. & Rec. & F1 & Prec. & Rec. & F1
       & Prec. & Rec. & F1 & Prec. & Rec. & F1 \\
\midrule
\multicolumn{13}{l}{\textit{MapTracker}~\citep{chen2024maptracker}} \\
VMA~\citep{VMA}       & 49.6 & 24.7 & 33.0 & 15.0 & 12.5 & 13.6 & {82.7} & 68.6 & \underline{75.0} & 24.4 & 21.8 & 23.0 \\
MonoLaM~\citep{MonoLaM}   & 52.6 & 48.5 & \underline{50.5} & 34.0 & 33.6 & \underline{33.8} & 69.9 & 80.1 & 74.7 & 38.1 & {35.0} & \underline{36.5} \\
\rowcolor{groupshade}
\textbf{Ours}& {54.4} & {48.7} & \textbf{51.4} & {42.8} & {41.0} & \textbf{41.9} & {74.7} & {82.1} & \textbf{78.2} & {48.1} & {43.3} & \textbf{45.5} \\
\midrule
\multicolumn{13}{l}{\textit{StreamMapNet}~\citep{StreamMapNet}} \\
VMA~\citep{VMA}     & 48.9 & 30.3 & 37.4 & 18.9 & 16.3 & 17.5 & {61.2} & 57.1 & 59.1 & 18.8 & 18.6 & 18.7 \\
MonoLaM~\citep{MonoLaM}   & {59.0} & {76.5} & \underline{66.6} & 37.7 & 39.5 & \underline{38.6} & 60.7 & {77.3} & \underline{68.0} & 32.8 & {33.3} & \underline{33.0} \\
\rowcolor{groupshade}
\textbf{Ours}& {68.9} & {76.4} & \textbf{72.4} & {46.8} & {47.7} & \textbf{47.2} & {64.5} & {78.7} & \textbf{70.9} & {40.8} & {37.9} & \textbf{39.3} \\
\bottomrule
\end{tabular}
\end{table}

\paragraph{Comparison with Map Aggregation Baselines.}
Table~\ref{tab:main_results} compares our method with VMA and MonoLaM using local predictions from MapTracker and StreamMapNet on Argoverse2 (AV2) and nuScenes. \ours consistently outperforms both baselines in F1 across all datasets, upstream detectors, and map categories. For instance, using StreamMapNet predictions on AV2, \ours improves F1 scores over VMA by 35.0 percentage points for lane dividers and 29.7 percentage points for road boundaries. Compared to the strongest baseline, MonoLaM, the corresponding improvements are 5.8 and 8.6 percentage points, respectively.
The improvement is particularly notable for road boundaries, which often have more complex shapes, including sharp curves and circular structures. Their local predictions can also contain fragmented, distorted, or duplicated segments.
These variations make it difficult to apply the fixed association and smoothing rules across different cases. By learning from diverse training examples, our approach can better handle these geometric structures and prediction errors.
\par
\begin{figure}[thb]
  \centering
  \includegraphics[width=1\textwidth]{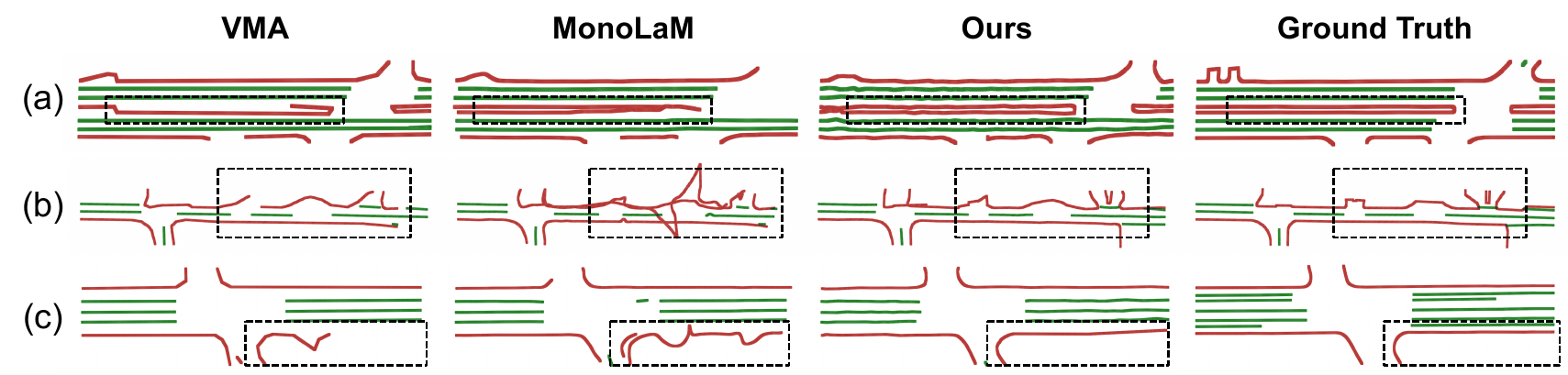}
  \caption{\small \textbf{Qualitative comparison with map aggregation baselines}. (a) StreamMapNet predictions on AV2, (b) StreamMapNet predictions on nuScenes, and (c) MapTracker predictions on AV2. Our \ours preserves road structures while reducing duplicate predictions and geometric artifacts.}
  \label{fig:main_exp_vis}
\end{figure}

We further evaluate how the methods handle these complex road structures and prediction errors through qualitative comparisons in Fig.~\ref{fig:main_exp_vis}.
In Fig.~\ref{fig:main_exp_vis}(a), a circular road boundary lies between several lane dividers. Both VMA and MonoLaM fail to recover the complete circular structure from the overlapping local predictions, while \ours reconstructs the complete circular geometry.
In Fig.~\ref{fig:main_exp_vis}(b), the highlighted road boundary is a single continuous line in the ground truth. VMA fails to merge fragments from different local predictions, leaving two disconnected segments with a large gap in between. MonoLaM produces several overlapping lines with distorted shapes. In contrast, our method recovers a single continuous boundary that more closely matches the ground truth.
Finally, Fig.~\ref{fig:main_exp_vis}(c) shows an example with substantial geometric noise along the road boundary. VMA removes some noisy fragments but also loses parts of the true boundary, while MonoLaM overfits the prediction noise and produces redundant lines. \ours reduces both noise and duplicate predictions while recovering a smoother boundary that is closer to the ground truth.

\paragraph{Ablation Studies.}
Table~\ref{tab:ablation_tokenization} studies the effects of coordinate tokenization, geometry-aware pretraining, and the line-level association loss using StreamMapNet predictions. \ours, highlighted in blue, achieves the highest F1 on both datasets and categories.
We first examine coordinate tokenization and pretraining. Compared with the vanilla tokenizer (Table~\ref{tab:ablation_tokenization}, row 1), coordinate tokenization with pretraining improves F1 across all four settings, particularly for lane dividers of AV2. Removing pretraining results in lower F1 scores in every setting (Table~\ref{tab:ablation_tokenization}, row 2). These results demonstrate the effectiveness of pretraining when introducing new coordinate tokens for map aggregation.
Moreover, the line-level association loss further improves F1 across all four settings compared with training only using the token-level generation loss (Table~\ref{tab:ablation_tokenization}, row 3). This suggests that explicit supervision over the line fragments helps the model associate related observations and reduce duplicate predictions.

\begin{table}[tbh]
\centering
\caption{\small \textbf{Ablation study on coordinate tokenization, coordinate-token pretraining, and the line-level contrastive objective}, using StreamMapNet predictions. Our proposed training strategy is highlighted in blue.}
\label{tab:ablation_tokenization}
\footnotesize
\setlength{\tabcolsep}{4.5pt}
\begin{tabular}{l cccccc cccccc}
\toprule
& \multicolumn{6}{c}{Argoverse2} & \multicolumn{6}{c}{nuScenes} \\
\cmidrule(lr){2-7}\cmidrule(lr){8-13}
& \multicolumn{3}{c}{Lane Divider} & \multicolumn{3}{c}{Road Boundary}
& \multicolumn{3}{c}{Lane Divider} & \multicolumn{3}{c}{Road Boundary} \\
\cmidrule(lr){2-4}\cmidrule(lr){5-7}\cmidrule(lr){8-10}\cmidrule(lr){11-13}
Setting & Prec. & Rec. & F1 & Prec. & Rec. & F1
        & Prec. & Rec. & F1 & Prec. & Rec. & F1 \\
\midrule
w/o Coord.\ Tok. & 54.6 & 38.7 & 45.3 & 41.1 & 38.6 & 39.8
                 & 62.0 & 77.2 & 68.7 & 35.7 & 32.3 & 33.9 \\
w/o Pretrain     & 63.4 & 69.3 & 66.2 & 41.8 & 41.2 & 41.5
                 & 60.8 & 74.9 & 67.1 & 38.8 & 32.8 & 35.5 \\
w/o Contrastive      & 67.3 & 73.5 & \underline{70.3} & 46.8 & 46.5 & \underline{46.6}
                 & 64.7 & 78.0 & \underline{70.7} & 40.6 & 36.2 & \underline{38.2} \\
\midrule
\rowcolor{groupshade}
\textbf{Ours (full)} & 68.9 & 76.4 & \textbf{72.4} & 46.8 & 47.7 & \textbf{47.2}
                     & 64.5 & 78.7 & \textbf{70.9} & 40.8 & 37.9 & \textbf{39.3} \\
\bottomrule
\end{tabular}
\end{table}
\section{Conclusion}

In this paper, we introduce \ours, a data-driven framework for vectorized map aggregation. 
Instead of relying on predefined association and smoothing rules, our approach formulates map aggregation as a conditional sequence generation problem given the local vectorized map observations.
To reduce dependence on specific detectors, we train on synthetic noisy local maps constructed from clean vector maps, where the noise simulates representative detection errors.
We further introduce coordinate tokenization with geometry-aware coordinate-token pretraining to better represent map elements, as well as a line-level association loss to improve association of map fragments. 
Experiments on Argoverse2 and nuScenes demonstrate that \ours consistently improves over heuristic and optimization-based aggregation baselines across multiple recent upstream detectors. 
Overall, our work highlights data-driven map aggregation as a powerful direction for reducing dependence on heuristic merging rules and improving the robustness of large-scale vectorized map construction.

\clearpage


\bibliography{iclr2027_conference}
\bibliographystyle{iclr2027_conference}


\clearpage
\appendix
\section{Appendix}

\begingroup
\etocsetnexttocdepth{subsubsection}
\etocsettocstyle
  {\bigskip
   \noindent{\large\bfseries Table of Contents}\par
   \smallskip\hrule\medskip}
  {\medskip\hrule\bigskip}
\localtableofcontents
\endgroup

\subsection{Implementation and Training Details}
\label{appendix:implementation_details}

\subsubsection{Training Setup}
All experiments are conducted on a single NVIDIA B200 GPU.
We use Qwen3.5-4B as the backbone and fine-tune it for map aggregation with LoRA, as described in Section~\ref{method:llm_map_aggregation}. We apply LoRA to the query and value projections in the attention layers, as well as the gate, up-projection, and down-projection layers in the feed-forward blocks.
We use a batch size of 1 with gradient accumulation over 8 steps. The learning rate is set to \(2\times10^{-4}\) with a cosine learning-rate schedule and 200 warm-up steps. We optimize the model using AdamW with a weight decay of 0.01. The checkpoint with the lowest validation loss is used for evaluation.


\subsubsection{Implementation Details}
We train separate models for Argoverse2 and nuScenes. For each dataset, a single aggregation model is trained on the synthetic local maps described in Section~\ref{sec:synthetic_data} and evaluated on predictions from both MapTracker and StreamMapNet, without detector-specific retraining.
\par
Before map serialization, we first translate all coordinates in the local observations and the target global map so that the ego position of the first frame becomes the origin:
\begin{equation}
    \mathbf{p}_i = \mathbf{p}_i^{\mathrm{global}} - \mathbf{o}_{T_1},
\end{equation}
where $\mathbf{o}_{T_1}\in\mathbb{R}^{2}$ denotes the ego position of the first frame in the original map coordinate system. Coordinates are represented at a resolution of $0.1\,\mathrm{m}$ over a range of $[-350,350]$ meters. The newly added coordinate-token rows in the input embedding matrix and LM head are initialized from the pretrained representations of the corresponding digit strings and rescaled to match the norms of the existing vocabulary representations.
\par
We serialize each map element as its category followed by its ordered point coordinates:
\begin{equation}
    (\mathbf{P},c) = \bigl[c,(x_1,y_1),\ldots,(x_N,y_N)\bigr],
\end{equation}
where $(x_i,y_i)$ denotes the translated coordinates of the $i$-th point. The coordinates are encoded using the new coordinate tokens. The input sequence contains local polylines from multiple frames, while the target sequence contains the corresponding clean global polylines. The numbers of frames, polylines, and points per polyline vary across different scenes, so the input and target sequences also vary in length.
\par
Since a global vectorized map is an unordered set of polylines, we use a deterministic ordering for the target polylines during training. Specifically, we first canonicalize the direction of each polyline and then sort the polylines according to their endpoint coordinates. This gives each target map a consistent serialization for autoregressive training. We also randomize the order of local polylines within each input frame during training to reduce sensitivity to input ordering.
\par
For the line-level association loss, we set both the loss weight $\lambda_{\mathrm{line}}$ and the temperature $\tau$ to 0.1.
\subsection{Pretraining Tasks for Coordinate Tokenization}
\label{appendix:pretraining_tasks}

We construct synthetic geometric and arithmetic tasks using 2D coordinates represented by the new coordinate tokens. The tasks include 2D coordinate translation, point-wise subtraction, midpoint prediction, point sorting, nearest-point selection, and bounding-box corner prediction.
Translation adds a displacement to a 2D point, while subtraction predicts the displacement between two points. Midpoint prediction computes the coordinate-wise average of two points, providing supervision for an operation relevant to merging nearby observations. Point sorting orders points first by $y$ and then by $x$, encouraging the model to learn a consistent spatial ordering. Nearest-point selection determines which of two candidate points is closer to a query point, providing supervision for geometric distance comparison. Bounding-box corner prediction returns the minimum and maximum coordinates of a polyline, capturing its spatial extent.
Each task is defined by an operation and its input coordinates, followed by the correct answer. We encode all coordinates using the new tokens and apply the training loss only to the answer tokens.
\par
We also pretrain the model on serialized local maps from the training scenes using next-token prediction. Given the current sequence prefix, the model predicts the next token in the serialized map sequence. This task is important, as it helps the model learn how the coordinate tokens are used within polyline sequences, which complements the numerical and spatial relationships learned from the geometric and arithmetic tasks. Since this pretraining task only involves local map observations, it does not require any ground-truth global maps.

\begin{table}[tbh]
\centering
\label{tab:pretrain_tasks}
\footnotesize
\setlength{\tabcolsep}{4pt}
\begin{tabular}{@{}lll@{}}
\toprule
\textbf{Task} & \textbf{Example} & \textbf{Downstream role} \\
\midrule
Translation   & \texttt{[10,2] + [-25,30] = [-15,32]}       & coordinate translation \\
Subtraction   & \texttt{[30,40] - [10,15] = [20,25]}        & offset between two points \\
Midpoint      & \texttt{mid: [10,20] [30,40] -> [20,30]}    & merging nearby observations \\
Sorting       & \texttt{sorty: [7,5] [2,1] -> [2,1] [7,5]}  & spatial ordering \\
Nearest point & \texttt{near [0,0]: [3,4] [10,2] -> [3,4]}  & geometric distance comparison \\
Bounding box  & \texttt{bbox: [3,9] [-4,2] -> [-4,2] [3,9]} & spatial extent of a polyline \\
\bottomrule
\end{tabular}
\caption{\small Synthetic pretraining tasks over the new coordinate tokens. The training loss is applied only to the answer tokens.}
\end{table}



\subsection{Training on Detector Outputs}

To evaluate whether training on the proposed synthetic dataset is sufficient for learning map aggregation, we compare our training strategy with directly training on real detector outputs. We keep the model architecture and two-stage training procedure unchanged and vary only the source of the noisy local maps used for training. The training set combines predictions from both MapTracker and StreamMapNet, and the resulting model is evaluated on real predictions from each detector.
\par
As shown in Table~\ref{tab:ablation_training_data}, training with synthetic local maps improves F1, particularly for road boundaries. For lane dividers, synthetic training also provides substantial gains on Argoverse2 and improves MapTracker on nuScenes, while the model trained on real detector outputs achieves slightly higher F1 on nuScenes lane dividers when evaluated on StreamMapNet predictions.
Overall, these additional results show that the proposed local map corruption process provides effective training signals and transfers well to different upstream map detectors.

\begin{table}[tbh]
\centering
\caption{\small \textbf{Training on synthetic data vs.\ real detector outputs.} Both rows use our method, and only the training datasets are different. Evaluation is on real MapTracker / StreamMapNet predictions.}
\label{tab:ablation_training_data}
\small
\setlength{\tabcolsep}{3.5pt}
\begin{tabular}{l cccccc cccccc}
\toprule
& \multicolumn{6}{c}{Argoverse2} & \multicolumn{6}{c}{nuScenes} \\
\cmidrule(lr){2-7}\cmidrule(lr){8-13}
& \multicolumn{3}{c}{Lane Divider} & \multicolumn{3}{c}{Road Boundary}
& \multicolumn{3}{c}{Lane Divider} & \multicolumn{3}{c}{Road Boundary} \\
\cmidrule(lr){2-4}\cmidrule(lr){5-7}\cmidrule(lr){8-10}\cmidrule(lr){11-13}
Training data
& Prec. & Rec. & F1 & Prec. & Rec. & F1
& Prec. & Rec. & F1 & Prec. & Rec. & F1 \\
\midrule
\rowcolor{groupshade}
\multicolumn{13}{l}{\textit{MapTracker}} \\
Detection outputs
& 51.6 & 37.1 & 43.1 & 33.7 & 31.5 & 32.6
& {80.7} & 73.5 & 76.9 & 42.4 & 31.9 & 36.4 \\
Synthetic (\textbf{Ours})
& {54.4} & {48.7} & \textbf{51.4} & {42.8} & {41.0} & \textbf{41.9}
& 74.7 & {82.1} & \textbf{78.2} & {48.1} & {43.3} & \textbf{45.5} \\
\midrule
\rowcolor{groupshade}
\multicolumn{13}{l}{\textit{StreamMapNet}} \\
Detection outputs
& 57.0 & 57.1 & 57.1 & 34.9 & 34.9 & 34.9
& {72.5} & 72.4 & \textbf{72.5} & 34.4 & 22.4 & 27.1 \\
Synthetic (\textbf{Ours})
& {68.9} & {76.4} & \textbf{72.4} & {46.8} & {47.7} & \textbf{47.2}
& 64.5 & {78.7} & 70.9 & {40.8} & {37.9} & \textbf{39.3} \\
\bottomrule
\end{tabular}
\end{table}




\subsection{Qualitative Results}

\subsubsection{Comparison with Map Aggregation Baselines}

\begin{figure}[thb]
  \centering
  \includegraphics[width=1\textwidth]{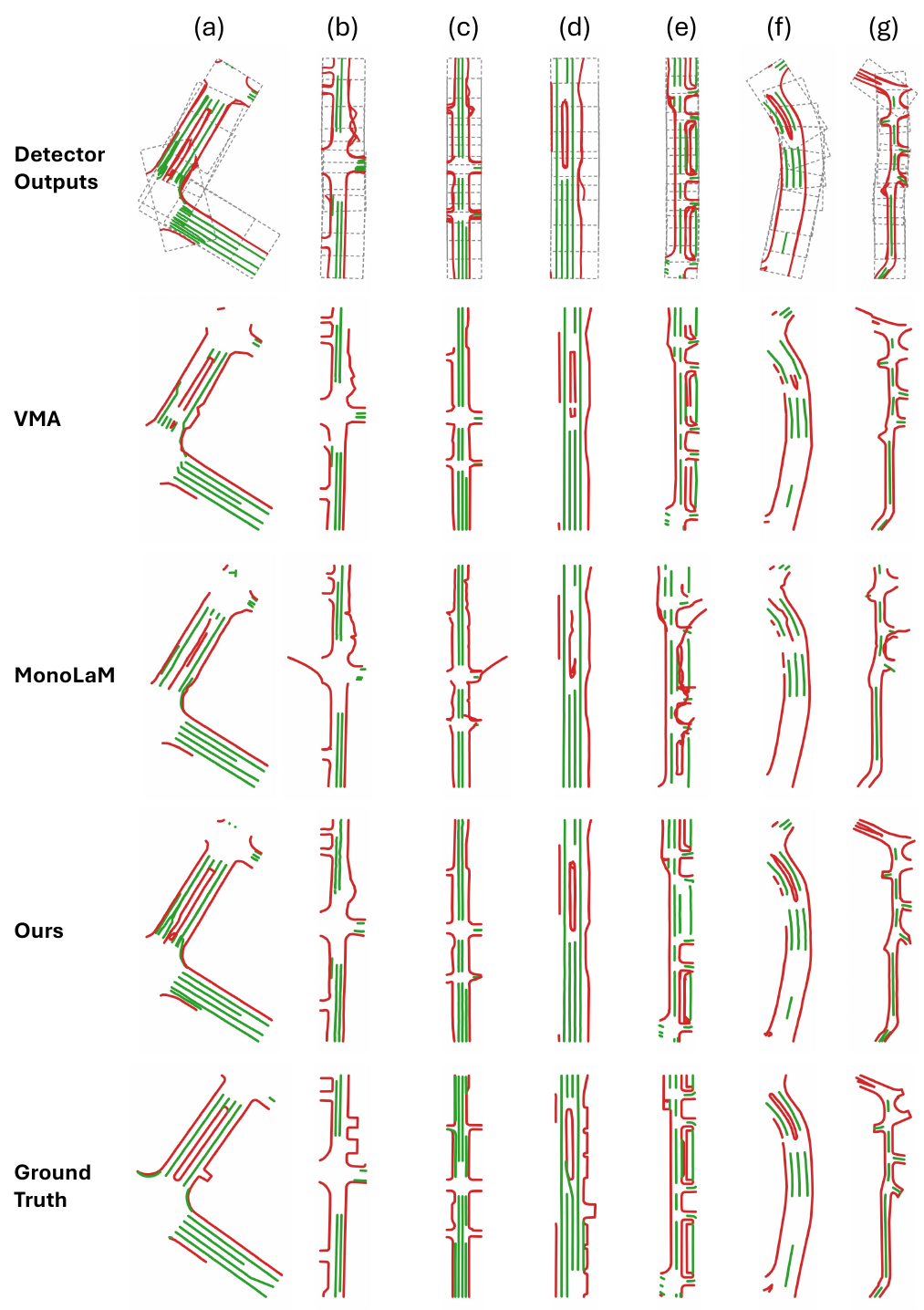}
  \caption{\small \textbf{Additional qualitative comparisons with map aggregation baselines.} (a)-(b) Argoverse2 with StreamMapNet predictions, (c)-(d) Argoverse2 with MapTracker predictions, (e) nuScenes with StreamMapNet predictions, and (f)-(g) nuScenes with MapTracker predictions.}
  \label{fig:app_comp_baselines}
\end{figure}

Fig.~\ref{fig:app_comp_baselines} presents additional qualitative comparisons with VMA~\citep{VMA} and MonoLaM~\citep{MonoLaM} on the Argoverse2 and nuScenes datasets using predictions from StreamMapNet~\citep{StreamMapNet} and MapTracker~\citep{chen2024maptracker}. These examples cover different road structures and aggregation challenges. 
VMA often fails to associate road-boundary fragments belonging to the same global map element, particularly near boundary endpoints in Fig.~\ref{fig:app_comp_baselines}(b) and (c) and along circular boundaries in Fig.~\ref{fig:app_comp_baselines}(d) and (f). These association errors result in fragmented or incomplete aggregated road boundaries. MonoLaM struggles to recover the underlying geometry of complex road boundaries, especially around circular structures or closely spaced map elements, as illustrated in Fig.~\ref{fig:app_comp_baselines}(a), (b), (e), and (f). It often introduces geometric distortions and redundant predictions in these regions. In comparison, \ours better associates related fragments and reconstructs road boundaries that more closely match the ground truth.

\subsubsection{Coordinate Tokenization and Pretraining}

\begin{figure}[thb]
  \centering
  \includegraphics[width=1\textwidth]{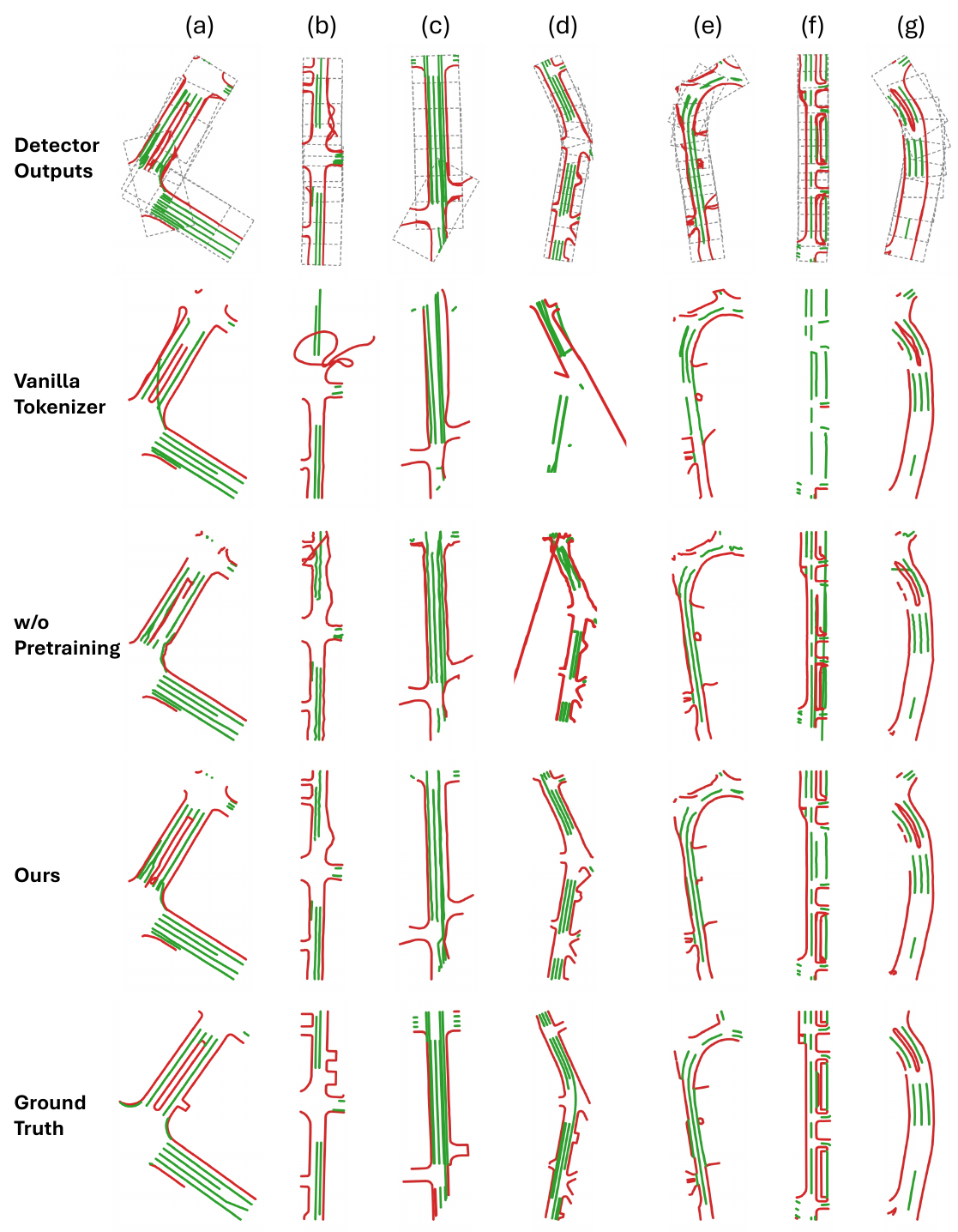}
  \caption{\small \textbf{Qualitative results on ablation study of coordinate tokenization and pretraining.} (a)-(c) Argoverse2 with StreamMapNet predictions, (d) Argoverse2 with MapTracker predictions, (e)-(f) nuScenes with StreamMapNet predictions, and (g) nuScenes with MapTracker predictions.}
  \label{fig:app_comp_tokenizers}
\end{figure}

Fig.~\ref{fig:app_comp_tokenizers} illustrates how coordinate tokenization and pretraining can affect the aggregated maps. The vanilla tokenizer often fails to generate road boundaries present in the ground truth, resulting in incomplete maps and lower recall. In addition, both vanilla tokenization and coordinate tokenization without pretraining can produce polylines that extend far beyond their intended endpoints, as illustrated by the long red lines in Fig.~\ref{fig:app_comp_tokenizers}(b) and (d). These errors suggest that these models struggle to determine where to stop generating a polyline. In contrast, our full model recovers more complete road structures and better preserves polyline endpoints, producing more complete and geometrically consistent maps.

\subsection{Further Discussion}

In the main paper, our experimental results demonstrate that vectorized map aggregation can be learned from synthetic local observations, which reduces the dependence on predefined association and merging rules. By combining the proposed coordinate-token pretraining with line-level association supervision, \ours improves aggregation across different upstream detectors on Argoverse2 and nuScenes datasets. These results also motivate further study of data-driven approaches for larger maps and other related applications.
\paragraph{Limitations and Future Work.}
Our experiments demonstrate effective aggregation over individual driving scenes. Extending this framework to longer driving trajectories is a natural direction for future work. However, this could be challenging because the growing number of local observations can exceed the context window of LLMs. Compared with standard subword tokenization, our coordinate tokenization already alleviates this limitation by representing each coordinate value as a single token, which allows more map observations to fit within the same context window. 
Incremental aggregation over overlapping regions is a promising future direction for handling longer driving trajectories while keeping each input sequence within the model's context window.
\par
The coordinate resolution of $0.1\mathrm{m}$ used in our experiments provides enough spatial detail and well supports the aggregation results reported on Argoverse2 and nuScenes. Adopting a finer resolution or extending the spatial range would further increase the vocabulary size. One possible direction is to explore hierarchical coordinate representations, which could accommodate larger maps or finer resolutions with fewer additional tokens.

\end{document}